\documentclass[conference]{IEEEtran}

\IEEEoverridecommandlockouts
\pdfoutput=1
\usepackage{hyperref}

\usepackage{cite}
\usepackage{svg}
\usepackage{amsmath,amssymb,amsfonts}
\usepackage{cleveref}
\usepackage{algorithmic}
\usepackage{subfigure}
\usepackage{graphicx}
\usepackage{textcomp}
\usepackage{xcolor}
\usepackage{CJKutf8}
\usepackage{array}
\usepackage{listings}
\usepackage[most,breakable]{tcolorbox}
\usepackage[numbers,sort]{natbib}
\def\BibTeX{{\rm B\kern-.05em{\sc i\kern-.025em b}\kern-.08em
    T\kern-.1667em\lower.7ex\hbox{E}\kern-.125emX}}
\begin{document}

\IEEEpubidadjcol

\title{Game-Oriented ASR Error Correction via RAG-Enhanced LLM\\

}

\author{
    \IEEEauthorblockN{Yan Jiang, Yongle Luo, Qixian Zhou, Elvis S. Liu$^{*}$\thanks{*Corresponding author}}
    \IEEEauthorblockA{Tencent Games}
    \IEEEauthorblockA{\{jocelyjiang, yongleluo, kaneqxzhou, elvissyliu\}@tencent.com}
}

\maketitle

\begin{abstract}
With the rapid development of the gaming industry and the increasing popularity of multiplayer online games, real-time voice communication has become a crucial tool for team collaboration and tactical exchanges. Automatic Speech Recognition (ASR) technology plays a vital role in modern gaming by converting voice commands into text, enabling efficient communication among players. However, existing general-purpose ASR systems face significant challenges in gaming scenarios due to the unique characteristics of in-game communication, such as short phrases, rapid speech, game-specific jargon, and environmental noise. These limitations often lead to frequent recognition errors, increasing communication costs and negatively impacting the overall gaming experience. Furthermore, the scarcity of domain-specific ASR training data exacerbates these issues, hindering system optimization.

To address the challenges of ASR systems in gaming scenarios, this study proposes the GO-AEC (Gaming-Oriented ASR Error Correction) framework. The framework leverages the generative capabilities of large language models (LLMs) and employs Retrieval-Augmented Generation (RAG) techniques with a game-specific knowledge base to better adapt to gaming environments. Additionally, we introduce a data augmentation strategy that combines LLMs with text-to-speech (TTS) techniques to enhance the diversity and robustness of game-specific datasets. The GO-AEC framework consists of three key modules: the data augmentation module, the N-best hypothesis-based LLM correction module, and the dynamic knowledge base module powered by RAG. Experimental results demonstrate that, compared to baseline methods, the proposed framework reduces the character error rate (CER) by 6.22\% and the sentence error rate (SER) by 29.71\%. These findings indicate that the GO-AEC framework effectively addresses the challenges of ASR error correction in gaming scenarios.
\end{abstract}

\begin{IEEEkeywords}
Large Language Model (LLM), ASR Error Correction, Data Augmentation, Retrieval-Augmented Generation (RAG)
\end{IEEEkeywords}
\section{Introduction}
In recent years, with the rapid development of the gaming industry and the widespread adoption of real-time voice communication, Automatic Speech Recognition (ASR) technology has become increasingly essential in gaming scenarios. Players rely on voice commands for collaboration, tactical communication, and interaction, making ASR systems a critical component of modern gaming experiences. However, compared to traditional ASR application domains, gaming scenarios present unique challenges and complexities.

First, gaming terminology evolves rapidly, with a vast array of domain-specific terms across different game genres (e.g., Multiplayer Online Battle Arena (MOBA), First-Person Shooter (FPS), Role-Playing Game (RPG)), which are often influenced by community trends and can change frequently. This dynamic nature of terminology makes it difficult for traditional ASR systems to effectively recognize and process such terms. Second, player voice inputs are often accompanied by background noise, accent variations, and informal language, further complicating the recognition process and posing higher requirements on the robustness of speech models. Lastly, the lack of training data in gaming scenarios limits the development and optimization of ASR models. Due to the significant domain differences across various game genres, existing ASR models lack transferability between games, resulting in poor performance in new gaming environments.

Although existing research has attempted to address these issues by developing domain-specific ASR models or optimizing post-processing modules, significant limitations remain. On one hand, training domain-specific ASR models requires large-scale, customized corpora, which are time-consuming and costly to collect for each game genre, making it impractical for real-world applications. On the other hand, current post-processing methods lack flexibility to adapt to the dynamic changes of terminology in gaming scenarios, leading to discrepancies between recognition results and actual semantic needs. Therefore, there is an urgent need for an ASR solution to address the core challenges in gaming scenarios.

To tackle these issues, considering that most ASR services used in gaming are black-box systems, this study proposes a novel Gaming-Oriented ASR Error Correction (GO-AEC) framework that is adaptable to different ASR services. The proposed framework is designed to address key challenges such as insufficient training data, noisy voice inputs, and dynamic terminology in gaming environments.  

First, to address the lack of training data across diverse gaming environments, we propose a hybrid data augmentation strategy based on Text-to-Speech (TTS), enabling the rapid generation of training datasets tailored to various gaming contexts.  
Second, we leverage the advanced contextual understanding and semantic inconsistency detection capabilities of large language models (LLMs). By designing prompt templates based on N-best hypotheses and applying supervised fine-tuning (SFT), we significantly enhance the ASR error correction accuracy in gaming environments, even under noisy input conditions.  
Finally, to ensure rapid adaptation to newly introduced terminology and effectively meet the evolving demands of gaming scenarios, we integrate a Retrieval-Augmented Generation (RAG)-based dynamic knowledge base mechanism. This mechanism dynamically retrieves relevant entries from domain-specific knowledge bases, enabling adaptability to new terms and concepts.

By adopting the aforementioned strategies, our framework has achieved significant improvements in ASR error correction performance within gaming scenarios. The main contributions of this study are summarized as follows:  
\begin{itemize}
\item We designed a hybrid data augmentation strategy that significantly enhances the system's generalization capability in data-scarce environments.
\item We proposed a novel ASR error correction framework that integrates language model fine-tuning with a dynamic knowledge base mechanism, achieving efficient performance optimization tailored to gaming scenarios.
\item We conducted a comprehensive and systematic evaluation of the proposed framework using the corpus from Tencent's game ``Arena Breakout''. Experimental results demonstrate that the framework exhibits significant advantages in reducing Character Error Rate (CER) and Sentence Error Rate (SER). Furthermore, we performed an in-depth analysis of the roles and effectiveness of each module within the framework, providing valuable insights for future improvements and applications.
\end{itemize}

\section{Related Work}
In this study, we primarily focus on post-processing error correction techniques for general ASR outputs. This section reviews the progress of related research in ASR error correction.

Early approaches to ASR error correction were predominantly based on system combination and statistical methods. A representative example is the ROVER system proposed by \citet{fiscus1997post}, which integrates outputs from multiple ASR systems using a voting mechanism to produce a composite result with reduced error rates. Building upon this, \citet{d2016automatic} introduced a machine translation-based paradigm that utilized phrase translation techniques to correct ASR n-best lists, achieving significant performance improvements.

With the advent of Transformer architectures, ASR error correction has seen substantial advancements. \citet{shin2019effective} demonstrated the effectiveness of BERT-based models in rescoring n-best hypotheses, leading to notable improvements in recognition accuracy. Similarly, \citet{li2021boost} combined BERT representations with a copy mechanism to retain correctly recognized tokens, effectively reducing word error rates. \citet{hrinchuk2020correction} proposed a Transformer-based encoder-decoder model that translates ASR outputs into grammatically correct text, achieving robust performance on noisy datasets. Further, \citet{dutta2022error} highlighted the utility of BART in ASR error correction by employing data augmentation during fine-tuning and leveraging alignment and rescoring strategies to reduce word error rates. Expanding on this, \citet{ma2023n} introduced an N-best T5 model that extracts richer information from the ASR decoding space to enhance transcription accuracy. Additionally, \citet{lee2024keyword} proposed a keyword-aware ASR error augmentation method that incorporates phonetic similarity errors, improving the robustness of dialogue state tracking in low-accuracy ASR environments.

Recently, the application of  LLMs has further advanced ASR error correction. For instance, \citet{song2023contextual} developed a contextual spelling correction system based on LLMs and prompt tuning, achieving significant reductions in word error rates. \citet{naderi2024towards} integrated confidence filtering with LLMs to improve transcription correction, particularly in low-performance ASR systems. Addressing the issue of over-correction, \citet{udagawa2024robust} proposed a conservative data filtering approach that screens low-quality training pairs, yielding improved accuracy in out-of-domain scenarios. \citet{ma2024asr} explored constrained decoding methods based on n-best lists and investigated the potential of zero-shot correction using LLMs. Furthermore, \citet{hu2024large} introduced a generative error correction framework that combines linguistic noise embeddings with knowledge distillation, enhancing correction performance in noisy environments.

In parallel, some researchers have incorporated RAG techniques into ASR error correction. \citet{li2024rag} proposed LA-RAG, which integrates token-level speech data storage with speech-to-speech retrieval mechanisms, significantly improving ASR accuracy. \citet{pusateri2024retrieval} developed a RAG-based strategy to address entity name errors in ASR systems, particularly for low-frequency entities. Extending this line of work, \citet{robatian2025gec} introduced the GEC-RAG model, which constructs a knowledge base pairing ASR predictions with ground truths. By employing TF-IDF to retrieve lexically similar examples, this model provides LLMs with relevant ASR error patterns, thereby improving correction accuracy.

Despite the remarkable progress in ASR error correction, research focusing on speech correction in gaming scenarios remains scarce. Gaming environments often face challenges such as limited training data, high background noise, and frequently updated terminologies, which have yet to be effectively addressed. In this paper, we propose a hybrid augmented data strategy that combines LLMs and TTS, specifically tailored for gaming scenarios. Furthermore, we integrate a RAG-based dynamic knowledge base mechanism with the capabilities of language models and explore fine-tuning strategies leveraging prompt and n-best hypotheses to enhance ASR error correction performance in gaming environments.

\section{Methodology}

\begin{figure}
\vskip 0.1in
    \centering
    \includegraphics[width=\linewidth]{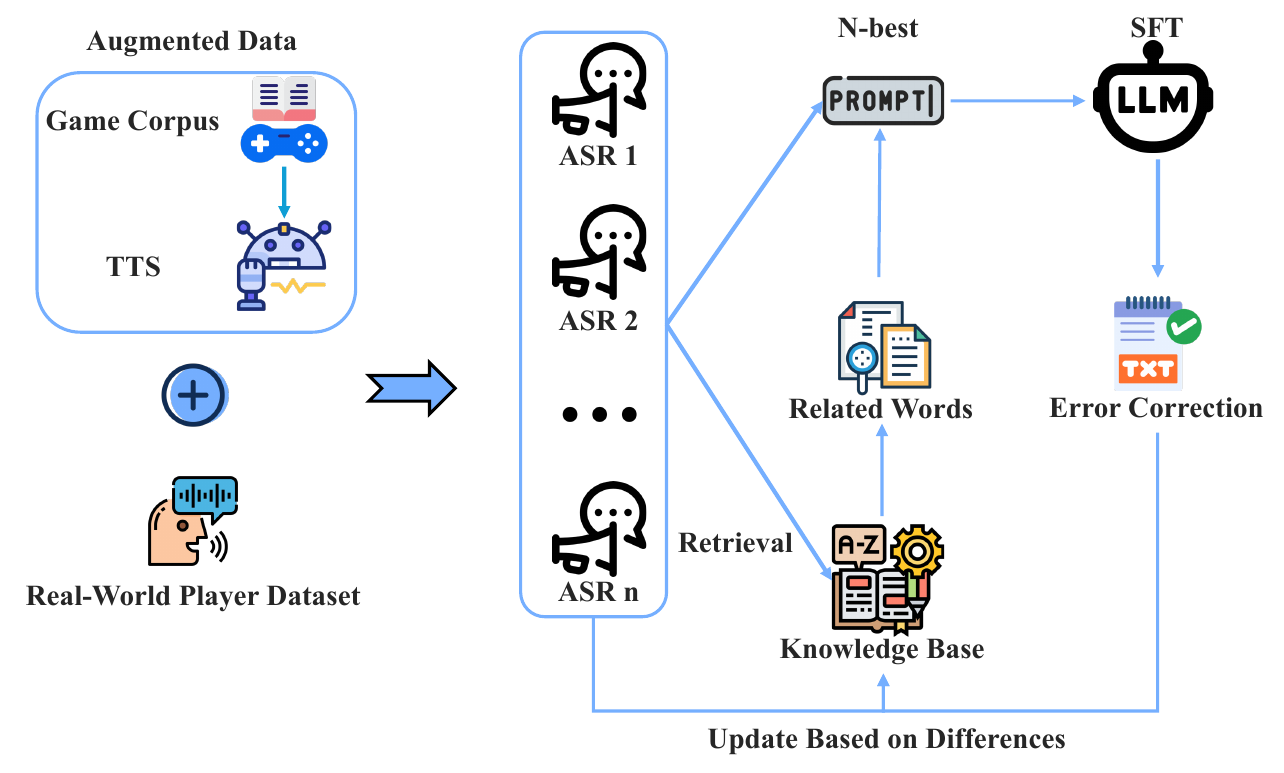}
    \caption{Gaming-Oriented ASR Error Correction Framework}
    \label{fig:model}
\vskip -0.2in
\end{figure}

This section provides an overview of the GO-AEC (Gaming-Oriented ASR Error Correction) framework and elaborates on its design and implementation. GO-AEC is specifically designed to address the core challenges of ASR error correction in gaming scenarios, including data scarcity, noise interference in voice inputs, and the dynamic nature of gaming terminology. As illustrated in Fig.~\ref{fig:model}, this section is divided into three parts, corresponding to the three key modules of the framework: the data augmentation module, the N-best hypothesis-based LLM correction module, and the retrieval-augmented generation module with a dynamic knowledge base. These modules work collaboratively to enhance the framework's ASR error correction capabilities, ensuring its effectiveness in addressing the complex demands of gaming scenarios.

\subsection{Data Augmentation Module}
Data scarcity is one of the primary challenges in ASR error correction for gaming scenarios. To address this issue, we proposed a hybrid data augmentation strategy that combines game-specific corpora, LLMs, and TTS technology to generate a diverse and comprehensive augmented dataset. This strategy involves three key steps:
\subsubsection{Domain-Specific Text Generation}
First, we extracted domain-specific terms and dialogue content from the game text corpus, denoted as the set $T = {t_1, t_2, \dots, t_n}$. These texts include common game-related terms, tactical commands, and player interaction content, such as ``retreat'', ``hide'', ``attack'', and ``resource gathering'' commands. To further enrich the diversity and coverage of the dataset, we utilized an LLM to expand and generate additional game-related text, resulting in an extended text set $T_{ext}$, as follows:
\begin{equation}
    T_{ext} = LLM(T).
\end{equation}
The final text dataset is obtained by merging the original text set and the extended text set, represented as:
\begin{equation}
    T_{final} = T \cup T_{ext}.
\end{equation}

\subsubsection{Speech Data Generation}
Using TTS technology, we converted the text data $T_{final}$ into corresponding audio data and designed a multi-dimensional speech generation strategy to simulate diverse acoustic characteristics and environmental conditions in real-world gaming scenarios. This strategy introduces variations in timbre, speed, background noise, and regional accents to reflect the complexity of gaming environments. The resulting speech dataset is denoted as $A = {a_1, a_2, \dots, a_n}$.

To emulate real gaming scenarios, we incorporated regional accent variations and diverse speech styles. For instance, rapid commands in emergencies feature a fast rate and urgent tone, while tactical discussions adopt a moderate pace and calm tone. By adjusting the TTS model's parameters, we generated speech data with varying styles, enhancing dataset diversity and ensuring alignment with real-world scenarios.

Additionally, we added background noises, including in-game sound effects and ambient sounds from different environments (e.g., train stations, shopping malls), to realistically simulate complex acoustic conditions. These enhancements ensure the dataset's realism and adaptability across various gaming scenarios and player environments.

By combining these strategies, we constructed a highly diverse and realistic speech dataset through multi-dimensional speech generation. This dataset provides rich training samples, significantly improving the model's generalization ability and robustness in complex gaming environments. The overall generation process can be expressed as:
\begin{equation}
A = \bigcup_{i=1}^n TTS \left( T_{final},v_i,n_i \right),
\end{equation}
where $v_i$ and $n_i$ represent the randomly selected voice profile and background noise parameters, respectively, $n$ denotes the total number of generated speech samples, and $\bigcup$ represents the union of all generated samples.

\subsubsection{Incorporation of Real Player Speech Data}
To further enhance the model's robustness to noise and accents in real-world environments, we incorporate a real player speech dataset, denoted as $R = {r_1, r_2, \dots, r_m}$.
The final augmented dataset is represented as:
\begin{equation}
    D_{aug} = A \cup R.
\end{equation}

By leveraging the hybrid augmentation strategy to generate the $D_{aug}$ dataset, we significantly improve the model's generalization capability across diverse gaming scenarios, enabling it to better adapt to complex gaming environments.

\subsection{N-best Hypothesis-based LLM Correction Module}
To enhance the error correction capability of ASR systems, we propose an N-best hypothesis-based language model correction module. This module takes the N-best hypothesis list as input, integrates game background information, and employs a SFT approach to optimize large language models for efficient correction of ASR outputs. Specifically, we extract candidate texts from the N-best hypothesis list $H = {h_1, h_2, \dots, h_k}$ generated by multiple ASR services and combine them with game background information $B$ to construct an input format suitable for the fine-tuned LLM. This design enables the SFT-optimized LLM to comprehensively evaluate multiple candidate texts to generate the optimal correction result.

To achieve this goal, we design an SFT framework that allows the LLM to learn from labeled data how to utilize the prompt structure for correction tasks. The design of the prompt is the core of this correction module. We use enumeration to clearly present the candidate texts from the N-best hypothesis list and randomize their order to avoid any bias toward a specific ASR system. The construction of the prompt can be formalized as:
\begin{equation}
    Prompt = Format(H,B),
\end{equation}
where $Format(\cdot)$ represents a formatting function that converts the candidate texts $H$ and background information $B$ into a natural language prompt. During training, the goal of supervised fine-tuning is to minimize a loss function that optimizes the model to generate semantically accurate results. The training objective can be formalized as:
\begin{equation}
    \mathcal{L} = \frac{1}{N} \sum _{i=1}^N Loss(h_i^*,h_i),
\end{equation}
where $h_i^* = LLM_{SFT}(Prompt_i)$ represents the predicted result generated by the model based on the $Prompt_i$; $h_i$ is the ground truth optimal text; $Loss(h_i^*, h_i)$ measures the difference between the predicted result $h_i^*$ and the ground truth $h_i$.

Specifically, the loss function $\text{Loss}(\cdot)$ can be defined as a token-level cross-entropy loss:
\begin{equation}
    Loss (h_i^*,h_i) = -\sum_{t=1}^T logP(h_{i,t}|h_{i<t},Prompt_i),
\end{equation}
where $T$ denotes the length of the target text $h_i$; $h_{i,t}$ represents the $t$-th token of the target text $h_i$; $P(h_{i,t} | h_{i<t}, Prompt_i)$ represents the probability of generating the target token $h_{i,t}$ given the $Prompt_i$ and the previously generated tokens $h_{i<t}$.

By minimizing this loss function, the model can better generate semantically accurate correction results.

\subsection{Retrieval-Augmented Generation Module with Dynamic Knowledge Base}
To accommodate the dynamic evolution of game-specific terminology, we have designed a RAG module based on a dynamic knowledge base. This module aims to enhance the correction capabilities of ASR systems, particularly in scenarios involving complex game terminology and dynamic contexts.

Specifically, we analyzed the differences between ASR outputs and ground truth texts in the training dataset, focusing on errors involving substitutions, insertions, or deletions of no more than $n$ characters. These differences were organized into a terminology knowledge base, which contains common erroneous words in the training set associated with specific terms, as well as phonetically similar words generated by the language model. The terminology knowledge base captures typical error patterns that ASR systems exhibit when processing dynamic terminology or complex scenarios.

During the construction of prompt templates, the RAG module dynamically retrieves whether the ASR results contain any erroneous words recorded in the terminology knowledge base. If matches are found, the module integrates the corresponding entries (term-error pairs) from the knowledge base as supplementary information into the prompt. This design enables the model to better understand the semantics and effectively leverage the dynamic knowledge base to generate more accurate correction results.

Furthermore, to meet the requirements of dynamic terminology updates, the module supports real-time updates to the terminology knowledge base. These updates are low-cost and plug-and-play, allowing the system to adapt to the evolving game terminology and linguistic expressions. The dynamically updated prompt template is as follows:

\begin{tcolorbox}[
  enhanced,
  title={\textbf{ASR Error Correction Prompt with RAG}},
  colback=white,
  colframe=blue!75!black,
  fonttitle=\bfseries\large,
  boxrule=0.5mm
]

\textbf{Role:} You are now an ASR error correction expert. Three ASR models have transcribed the user's speech input into text. Your task is to correct the ASR transcriptions based on the outputs of these three models.

\textbf{Conversation Context:} 

[Relevant game background information]

\tcbline

\textbf{Terminology Knowledge Base:}

correct word $|$ erroneous word

\tcbline

\textbf{Output Requirements:}
\begin{enumerate}
  \item Please correct the ASR text based on terminology knowledge base and output \{Corrected ASR Text\}.
  \item If you determine that no correction is needed, directly output \{Original Text\}.
  \item Do not output the content of the instruction template; only output the specified result.
  \item If the text is abnormal but the terminology knowledge base does not contain relevant vocabulary, you need to judge independently. If you cannot correct it, output \{Original Text\}.
\end{enumerate}

\textbf{Tips:} You need to directly correct obvious abnormal keywords and output the corrected ASR text.

\tcbline

\textbf{Let's begin:}

ASR 1 Output: [req1]

ASR 2 Output: [req2]

ASR 3 Output: [req3]

\textbf{Your Output:}
\end{tcolorbox}

\subsection{Inference}
During the inference phase, the GO-AEC framework leverages a fine-tuned LLM to provide real-time ASR error correction, ensuring efficiency and adaptability in dynamic gaming environments. 

The process begins with multiple ASR systems generating N-best hypothesis lists ($H$), which capture various possible transcriptions of the user's speech input. These hypotheses are then combined with relevant game background information ($B$) to construct dynamic prompts. Simultaneously, the RAG module dynamically queries the knowledge base to retrieve entries related to known errors. If relevant matches are found, they are integrated into the prompt as supplementary information to further enhance correction accuracy.

Once the dynamic prompt is constructed, the fine-tuned LLM evaluates the N-best hypotheses and generates the most accurate correction based on the provided context and knowledge base. The corrected text is then returned as the final output, ready to be utilized in downstream tasks such as executing in-game commands or facilitating player interactions.

To maintain adaptability in evolving gaming environments, the GO-AEC framework incorporates a real-time knowledge base update mechanism. When new terminology or error patterns are encountered during inference, they are logged and subsequently used to update the knowledge base. This mechanism ensures that the system can swiftly adapt to changes in gaming terminology and player behavior, providing a scalable, robust, and efficient solution for ASR error correction in complex and dynamic environments.

\section{Experiments}
This section details the experimental setup, including datasets, evaluation metrics, and baseline methods. Additionally, we report the experimental results and provide an in-depth analysis through ablation experiments and case studies.

\subsection{Experimental Setup}
\subsubsection{Datasets}

To validate the effectiveness of the GO-AEC model, we constructed a hybrid dataset combining TTS-based synthetic data with real-world player speech data collected from gaming test scenarios (with informed consent). The experiments focus on Chinese ASR tasks, using the first-person shooter game \textbf{Arena Breakout} as a case study. This game features tactical objectives like ``search'', ``attack'', and ``retreat'', with domain-specific terminology, rapid commands, and complex noisy environments (e.g., gunfire, explosions, footsteps), making it an ideal benchmark for evaluating ASR error correction systems in terms of language understanding and noise robustness.

The hybrid dataset comprises two components:
\begin{enumerate}
    \item Synthetic Speech Data: Using TTS systems, we generated speech samples simulating the linguistic and acoustic diversity of Arena Breakout. A curated Chinese text corpus covering tactical actions (e.g., retreat, scout, attack, gather supplies) was used to create diverse samples with regional accents, varied speech rates, and environmental noise (e.g., gunfire, footsteps, explosions).
    \item Real-World Player Speech Data: Speech data was collected from test players during gameplay, capturing authentic tactical commands and situational updates. Recordings included in-game sounds, environmental noise, microphone distortions, and overlapping speech. All recordings were manually transcribed to ensure high-quality annotations for model training and evaluation.
\end{enumerate}

This dataset realistically reflects the challenges of gaming scenarios, providing a robust foundation for assessing GO-AEC's performance.

\begin{table}[ht]
    \centering
    \caption{Dataset Statistics}
    \renewcommand{\arraystretch}{1.2} 
    \setlength{\tabcolsep}{4pt} 
    \begin{tabular}{p{2.5cm}|p{1.5cm}|p{1.5cm}|p{1.5cm}}
        \hline
        \textbf{Dataset} & \textbf{Training} & \textbf{Validation} & \textbf{Test} \\ \hline
        \textbf{Number of Samples}  & 32,754 & 3,579 & 3,872 \\ 
        \textbf{Total Duration (hrs)} & 22.16 & 1.98 & 2.58 \\
        \textbf{Source}  & TTS; Player & Player & Player \\ \hline
    \end{tabular}
    \label{tab:dataset_statistics}
\end{table}

To ensure scientific rigor, the dataset was divided into training, validation, and test sets. The training set consisted of both synthetic and real-world speech data to enhance the model's generalization capabilities and address the issue of data scarcity, while the validation and test sets contained only real-world data to evaluate the model's performance in realistic scenarios. Table~\ref{tab:dataset_statistics} presents the dataset statistics, and Table~\ref{tab:dataset_examples} provides representative examples.

\begin{CJK}{UTF8}{gbsn}
\begin{table}[ht]
    \centering
    \caption{Dataset Examples}
    \begin{tabular}{m{4cm}|m{1.3cm}|m{1.5cm}}
        \hline
        \centering \textbf{Text} & \textbf{Speech Source} & \textbf{Background Noise} \\ \hline
        \centering 给我一个绷带 \\ \textit{Give me a bandage.} &  TTS & Footsteps \\ \hline
        \centering 前往撤离点 \\ \textit{Head to the exfil.} & Player Speech & None \\ \hline
        \centering 帮我舔包 \\ \textit{Check the loot for me.} &  TTS & Gunfire \\ \hline
    \end{tabular}
    \label{tab:dataset_examples}
\end{table}
\end{CJK}

\subsubsection{Evaluation Metrics}
To quantitatively evaluate the performance of ASR error correction, we employed the following metrics:

\begin{itemize}
    \item Character Error Rate (CER): Given that the dataset is in Chinese, we adopt CER as the evaluation metric. It serves as an effective metric for evaluating the performance of ASR error correction systems in Chinese contexts. CER is defined as:
    \begin{equation}
        CER = \frac{S+D+I}{N} \times 100,
    \end{equation}
    where $S$, $D$, and $I$ represent the number of substitutions, deletions, and insertions, respectively, and $N$ is the total number of characters in the reference transcript.
    \item Sentence Error Rate (SER): While CER focuses on character-level errors, SER measures the percentage of sentences with at least one error, providing a stricter evaluation criterion. SER is defined as:  
    \begin{equation}
        SER = \frac{E}{T} \times 100,
    \end{equation}
    where $E$ represents the number of sentences containing at least one character error, and $T$ is the total number of sentences. SER is particularly useful in scenarios like gaming commands, where even a single error in a sentence can lead to incorrect actions.  
\end{itemize}

\subsubsection{Implementation Details}
The GO-AEC framework is implemented using PyTorch. Considering the high performance and real-time requirements in gaming scenarios, we adopted the lightweight language model Qwen2.5-1.5B as the base model. All experiments were conducted on an NVIDIA A100 GPU, with the learning rate set to $1 \times 10^{-5}$ and a batch size of $32$. For synthetic data, we randomly selected from 6 voice tones and 238 noise source, while also randomly adjusting volume and speech speed to enhance the model's robustness. For the N-best hypothesis set, ASR results for the same speech input were generated by multiple mainstream ASR cloud services (e.g., Tencent, Alibaba, ByteDance). The GO-AEC model effectively utilizes these candidate results to perform error correction, thereby significantly improving correction performance.

\subsection{Baseline Methods}
To comprehensively evaluate the performance of our proposed model GO-AEC, we implemented and compared it against the following baseline methods. Since there has been no prior work specifically addressing ASR error correction in gaming scenarios, we developed these baselines for a fair comparison:
\begin{itemize}
    \item Vanilla ASR Output: The raw output from the best ASR system without any post-processing, used to evaluate the error level in the original ASR output.
    \item T5-Based: T5 is trained on gaming-specific data to assess its effectiveness in correcting ASR errors within gaming scenarios. 
    \item BART-Based: BART is trained on gaming-specific data to assess its effectiveness in correcting ASR errors within gaming scenarios.  
    \item LLM w/o SFT:  A general-purpose LLM (Qwen2.5-72B) used for ASR error correction without domain-specific fine-tuning, generating corrections using the same prompts as our framework.
    \item LLM w/ SFT: The same LLM (Qwen2.5-1.5B) fine-tuned on gaming-specific ASR outputs and ground truth transcripts. This baseline uses the same prompts as our proposed framework but does not integrate RAG.  
    \item GO-AEC: Our proposed framework integrates multiple components tailored for gaming speech correction, N-best hypotheses to improve robustness, and a RAG module with a dynamic knowledge base for handling domain-specific vocabulary. 
\end{itemize}

\subsection{Results}
Table~\ref{tab:result} presents the comparison of GO-AEC with baseline methods in terms of Character Error Rate (CER) and Sentence Error Rate (SER). GO-AEC achieves a 6.22\% reduction in CER and a 29.71\% reduction in SER compared to the vanilla ASR output, significantly outperforming all baselines and demonstrating its superior performance in gaming ASR error correction.

Compared to traditional sequence-to-sequence models such as T5 and BART, GO-AEC exhibits stronger domain adaptability. While T5 and BART improve the ASR output to some extent, their limited ability to capture domain-specific linguistic patterns constrains their performance.

Large language models without fine-tuning demonstrate general-purpose error correction capabilities, while the fine-tuned Qwen-1.5B further enhances performance, highlighting the importance of domain-specific fine-tuning. 
As shown in the Table~\ref{tab:result}, GO-AEC significantly outperforms the untuned Qwen-72B (Qwen-72B w/o SFT) in both the CER and SER metrics. Moreover, Qwen-72B comprises 48 times as many parameters as GO-AEC.
By integrating the RAG module and N-best hypotheses, GO-AEC further optimizes CER and SER compared to the simple supervised fine-tuning of the qwen-1.5B (Qwen-1.5B w/ SFT)), showcasing its remarkable advantages in handling complex domain-specific contexts and linguistic structures.

\begin{table}[ht]
    \centering
    \caption{Performance Comparison of GO-AEC and Baseline Methods}
    \begin{tabular}{c|c|c}
        \hline
        \textbf{Method} & \textbf{CER} & \textbf{SER} \\ \hline
        Vanilla ASR Output & 9.87\% & 46.51\% \\
        % Regex Matching & & \\
        T5-Based & 9.60\% & 45.51\% \\
        Bart-Based & 9.22\% & 45.06\% \\
        Qwen-72B w/o SFT & 7.07\% & 34.68\%\\
        Qwen-1.5B w/ SFT & 6.79\% & 24.66\% \\
        \textbf{GO-AEC (ours)} & \textbf{3.65\%} & \textbf{16.80\%} \\ \hline
    \end{tabular}
    \label{tab:result}
\end{table}
\subsection{Analysis}

To further understand the advantages of GO-AEC, we conducted a detailed analysis from multiple perspectives, including ablation experiments and qualitative case studies. These analyses provide insights into the contributions of individual components of our framework and highlight its strengths in real-world applications.

\subsubsection{Ablation Experiment}
To comprehensively evaluate the performance of the GO-AEC framework, we conducted ablation experiments to analyze the contributions of its components, including removing the RAG module (GO-AEC w/o RAG), excluding N-best hypotheses (GO-AEC w/o N-best), and using only pre-trained models without domain-specific fine-tuning (GO-AEC w/o SFT). The experimental results, as shown in Table~\ref{tab:ablation}, illustrate the performance variations under different configurations.

\begin{table}[ht]
    \centering
    \caption{Ablation Experiment Results for GO-AEC}
    \begin{tabular}{c|c|c}
        \hline
        \textbf{Method} & \textbf{CER} & \textbf{SER} \\ \hline
        \textbf{GO-AEC (ours)} & \textbf{3.65\%} & \textbf{16.80\%} \\
        GO-AEC w/o RAG & 6.79\% & 24.66\% \\ 
        GO-AEC w/o N-best & 6.84\% & 23.58\% \\
        GO-AEC w/o SFT  & 13.66\% & 50.10\% \\ \hline
    \end{tabular}
    \label{tab:ablation}
\end{table}

Specifically, removing the RAG module results in a significant decline in the model's ability to handle game-specific terms and expressions. This highlights that the retrieval-augmented mechanism provides essential domain knowledge, contributing to more precise semantic understanding and correction. Excluding the N-best hypotheses reduces the model's robustness in handling ambiguities in speech recognition, demonstrating that leveraging multiple candidate outputs effectively enhances correction accuracy. The most significant performance degradation occurs when domain-specific fine-tuning is removed. Even with the RAG module and N-best hypotheses retained, the lack of fine-tuning for the gaming context severely limits the model's understanding and correction capabilities in game-specific scenarios. 

In summary, the experimental results clearly demonstrate that the synergy of the RAG module, N-best hypotheses, and domain-specific fine-tuning enables GO-AEC to achieve efficient and accurate correction in complex gaming speech scenarios, showcasing its potential in real-world applications.

\subsubsection{Impact of N-best Hypotheses}
To evaluate the impact of N-best hypotheses, we generated N-best hypothesis sets using three different ASR cloud services (ByteDance, labeled as ASR-B; Alibaba, labeled as ASR-A; and Tencent, labeled as ASR-T) and compared these results with those obtained by using the recognition output of a single ASR service.

\begin{table}[ht]
    \centering
    \caption{Experimental Results for Different N Values}
    \begin{tabular}{c|c|c}
        \hline
        \textbf \textbf{Data Source} & \textbf{CER} & \textbf{SER} \\ \hline
         ASR-B & 8.94\% & 28.74\% \\
         ASR-A & 7.13\% & 23.11\% \\
         ASR-T & 6.84\% & 23.58\% \\
         \textbf{GO-AEC (ours)} & \textbf{3.65\%} & \textbf{16.80\%} \\ \hline
    \end{tabular}
    \label{tab:nbest_analysis}
\end{table}

The experimental results, summarized in Table~\ref{tab:nbest_analysis}, show that leveraging the complementary strengths of multiple ASR services significantly improves error correction performance. When relying on a single ASR service, even the best-performing ASR-T struggles to meet practical application requirements in terms of CER and SER. Although there are performance differences among the ASR services, even the relatively weaker ones contribute valuable insights to the overall error correction process. By integrating the outputs of all three ASR services, the CER and SER of GO-AEC are significantly reduced.

This study demonstrates that combining the outputs of multiple ASR systems allows GO-AEC to more effectively capture potential correct transcriptions, thereby enhancing both accuracy and robustness. This approach is particularly advantageous in gaming scenarios, where speech often includes specialized terminology and rapid commands that a single ASR system may struggle to recognize accurately.

\subsubsection{Effect of Fine-Tuning with Different Data Sizes}
To evaluate the effect of fine-tuning with varying amounts of gaming-specific data, we conducted experiments using subsets of the dataset with different data sizes. Specifically, we selected 10\%, 25\%, 50\%, 75\%, and 100\% of the total available data for fine-tuning. The performance of the model was measured in terms of CER and SER, as presented in Table~\ref{tab:finetuning_analysis}.

The experimental results on fine-tuning data size demonstrate the impact of data scale on model performance. As the training data increases, the model's performance generally improves. In particular, during the increase of training data from 10\% to 25\%, the CER significantly drops from 5.95\% to 4.13\%, while the SER decreases from 23.00\% to 20.00\%.

Notably, even with only 10\% of the training data, GO-AEC achieves relatively good performance (with CER at 5.95\%), indicating the model's strong data efficiency. When the data reaches 50\%, the performance is already close to that achieved using the full dataset, suggesting that a medium-scale dataset is sufficient for the model to learn most of the useful patterns. This outcome holds significant practical implications, as it indicates that under resource constraints, employing a smaller yet high-quality dataset can yield near-optimal performance.

\begin{table}[ht]
    \centering
    \caption{Performance of Fine-Tuning with Different Data Sizes}
    \begin{tabular}{c|c|c}
        \hline
         \textbf{Data Size} & \textbf{CER} & \textbf{SER} \\ \hline
         10\% & 5.95\% & 23.00\% \\
         25\% & 4.13\% & 20.00\% \\
         50\% & 4.38\% & 19.63\% \\
         75\% & 4.04\% & 18.26\% \\
         100\% & \textbf{3.65\%} & \textbf{16.80\%} \\ \hline
    \end{tabular}
    \label{tab:finetuning_analysis}
\end{table}

\subsubsection{Case Study}
We further conducted a qualitative analysis through case studies to illustrate the effectiveness of GO-AEC in real-world gaming scenarios. Table~\ref{tab:casestudy} presents several representative examples comparing the inputs and outputs of GO-AEC. The examples highlight GO-AEC's remarkable ability to correct ASR output errors and adapt them to the specific context of gaming. By leveraging contextual information and gaming-specific semantics, GO-AEC effectively identifies and rectifies common recognition issues in ASR systems. Whether addressing lexical confusion caused by phonetic similarity or refining generic outputs into more targeted and actionable commands, GO-AEC demonstrates a strong capacity for semantic understanding and contextual adaptation.
\begin{CJK}{UTF8}{gbsn}
\begin{table}[ht]
    \centering
    \caption{Case Study Examples}
    \begin{tabular}{c|c}
        \hline
        \textbf{ASR 1 Output}  & \textbf{GO-AEC} \\ \hline
        哪里\textcolor{red}{预习}了 & 哪里\textcolor{blue}{遇袭}了 \\
        \multicolumn{1}{c|}{\textit{Where to preview}} & \multicolumn{1}{c}{\textit{Where was attacked}} \\ \hline
        \textcolor{red}{DNA}在哪 & \textcolor{blue}{敌人}在哪儿 \\
        \multicolumn{1}{c|}{\textit{Where is the DNA}} & \multicolumn{1}{c}{\textit{Where is the enemy}} \\ \hline
        \textcolor{red}{适合}去救一下 & \textcolor{blue}{4号}去救一下 \\
        \multicolumn{1}{c|}{\textit{Suitable to rescue}} & \multicolumn{1}{c}{\textit{No.4 go to rescue}} \\ \hline
    \end{tabular}
    \label{tab:casestudy}
\end{table}
\end{CJK}

These examples illustrate that GO-AEC not only significantly improves the accuracy of speech recognition but also enhances the relevance and practicality of semantic outputs. Particularly in gaming scenarios, GO-AEC excels at transforming vague or incorrect commands into clear and precise operational instructions, making it highly effective in environments characterized by rapid commands and specialized terminology.

\section{Conclusion}
We proposed a Gaming-Oriented ASR Error Correction (GO-AEC) framework to address the core challenges faced by current ASR systems in gaming scenarios, including dynamic data scarcity, high-noise environments, and terminology. By designing a hybrid data augmentation strategy, leveraging the contextual understanding capabilities of LLMs, and introducing a RAG-based dynamic knowledge base mechanism, the GO-AEC framework achieved significant improvements in multiple aspects.

Experimental results demonstrated that the GO-AEC framework exhibits notable advantages in reducing CER and SER, particularly in complex gaming speech scenarios. Furthermore, detailed analysis of each module within the framework validated the critical roles of the hybrid data augmentation strategy, language model fine-tuning, and the dynamic knowledge base mechanism in enhancing ASR error correction performance.

In summary, this study not only provides an effective solution to the ASR challenges in gaming scenarios but also offers new insights for developing more robust ASR error correction systems in other domains. However, further exploration is needed in the following directions: first, enhancing the framework's adaptability to multilingual conditions to support a broader range of global gaming scenarios; second, optimizing the retrieval and generation efficiency of the dynamic knowledge base to meet the speed requirements of real-time voice interactions; and third, extending the framework's applicability to more gaming scenarios, such as voice interactions with AI teammates. Future work will focus on these directions to further refine the GO-AEC framework and explore its potential applications in other real-time voice interaction scenarios.

\bibliographystyle{unsrtnat}
\bibliography{mybib}

\begin{thebibliography}{16}
\providecommand{\natexlab}[1]{#1}
\providecommand{\url}[1]{\texttt{#1}}
\expandafter\ifx\csname urlstyle\endcsname\relax
  \providecommand{\doi}[1]{doi: #1}\else
  \providecommand{\doi}{doi: \begingroup \urlstyle{rm}\Url}\fi

\bibitem[Fiscus(1997)]{fiscus1997post}
Jonathan~G Fiscus.
\newblock A post-processing system to yield reduced word error rates: Recognizer output voting error reduction (rover).
\newblock In \emph{1997 IEEE workshop on automatic speech recognition and understanding proceedings}, pages 347--354. IEEE, 1997.

\bibitem[D’Haro and Banchs(2016)]{d2016automatic}
Luis~Fernando D’Haro and Rafael~E Banchs.
\newblock Automatic correction of asr outputs by using machine translation.
\newblock In \emph{Interspeech}, volume 2016, pages 3469--3473, 2016.

\bibitem[Shin et~al.(2019)Shin, Lee, and Jung]{shin2019effective}
Joongbo Shin, Yoonhyung Lee, and Kyomin Jung.
\newblock Effective sentence scoring method using bidirectional language model for speech recognition.
\newblock \emph{arXiv preprint arXiv:1905.06655}, 2019.

\bibitem[Li et~al.(2021)Li, Di, Wang, Ouchi, and Lu]{li2021boost}
Wenkun Li, Hui Di, Lina Wang, Kazushige Ouchi, and Jing Lu.
\newblock Boost transformer with bert and copying mechanism for asr error correction.
\newblock In \emph{2021 International Joint Conference on Neural Networks (IJCNN)}, pages 1--6. IEEE, 2021.

\bibitem[Hrinchuk et~al.(2020)Hrinchuk, Popova, and Ginsburg]{hrinchuk2020correction}
Oleksii Hrinchuk, Mariya Popova, and Boris Ginsburg.
\newblock Correction of automatic speech recognition with transformer sequence-to-sequence model.
\newblock In \emph{Icassp 2020-2020 ieee international conference on acoustics, speech and signal processing (icassp)}, pages 7074--7078. IEEE, 2020.

\bibitem[Dutta et~al.(2022)Dutta, Jain, Maheshwari, Pal, Ramakrishnan, and Jyothi]{dutta2022error}
Samrat Dutta, Shreyansh Jain, Ayush Maheshwari, Souvik Pal, Ganesh Ramakrishnan, and Preethi Jyothi.
\newblock Error correction in asr using sequence-to-sequence models.
\newblock \emph{arXiv preprint arXiv:2202.01157}, 2022.

\bibitem[Ma et~al.(2023)Ma, Gales, Knill, and Qian]{ma2023n}
Rao Ma, Mark~JF Gales, Kate~M Knill, and Mengjie Qian.
\newblock N-best t5: Robust asr error correction using multiple input hypotheses and constrained decoding space.
\newblock In \emph{INTERSPEECH}, 2023.

\bibitem[Lee et~al.(2024)Lee, Im, Lee, and Lee]{lee2024keyword}
Jihyun Lee, Solee Im, Wonjun Lee, and Gary~Geunbae Lee.
\newblock Keyword-aware asr error augmentation for robust dialogue state tracking.
\newblock \emph{arXiv preprint arXiv:2409.06263}, 2024.

\bibitem[Song et~al.(2023)Song, Wu, Pundak, Chandorkar, Joshi, Velez, Caseiro, Haynor, Wang, Siddhartha, et~al.]{song2023contextual}
Gan Song, Zelin Wu, Golan Pundak, Angad Chandorkar, Kandarp Joshi, Xavier Velez, Diamantino Caseiro, Ben Haynor, Weiran Wang, Nikhil Siddhartha, et~al.
\newblock Contextual spelling correction with large language models.
\newblock In \emph{2023 IEEE Automatic Speech Recognition and Understanding Workshop (ASRU)}, pages 1--8. IEEE, 2023.

\bibitem[Naderi et~al.(2024)Naderi, Hermann, Nanchen, Hovsepyan, et~al.]{naderi2024towards}
Maryam Naderi, Enno Hermann, Alexandre Nanchen, Sevada Hovsepyan, et~al.
\newblock Towards interfacing large language models with asr systems using confidence measures and prompting.
\newblock In \emph{Proc. Interspeech 2024}, pages 2980--2984, 2024.

\bibitem[Udagawa et~al.(2024)Udagawa, Suzuki, Muraoka, and Kurata]{udagawa2024robust}
Takuma Udagawa, Masayuki Suzuki, Masayasu Muraoka, and Gakuto Kurata.
\newblock Robust asr error correction with conservative data filtering.
\newblock In \emph{Proceedings of the 2024 Conference on Empirical Methods in Natural Language Processing: Industry Track}, pages 256--266, 2024.

\bibitem[Ma et~al.(2024)Ma, Qian, Gales, and Knill]{ma2024asr}
Rao Ma, Mengjie Qian, Mark Gales, and Kate Knill.
\newblock Asr error correction using large language models.
\newblock \emph{arXiv preprint arXiv:2409.09554}, 2024.

\bibitem[Hu et~al.(2024)Hu, Chen, Yang, Li, Zhang, Chen, and Chng]{hu2024large}
Yuchen Hu, Chen Chen, Chao-han~Huck Yang, Ruizhe Li, Chao Zhang, Pin-Yu Chen, and Ensiong Chng.
\newblock Large language models are efficient learners of noise-robust speech recognition.
\newblock In \emph{International Conference on Learning Representations}, 2024.

\bibitem[Li et~al.(2024)Li, Shang, Wei, Guo, Li, He, Zhang, and Yang]{li2024rag}
Shaojun Li, Hengchao Shang, Daimeng Wei, Jiaxin Guo, Zongyao Li, Xianghui He, Min Zhang, and Hao Yang.
\newblock La-rag: Enhancing llm-based asr accuracy with retrieval-augmented generation.
\newblock \emph{arXiv preprint arXiv:2409.08597}, 2024.

\bibitem[Pusateri et~al.(2024)Pusateri, Walia, Kashi, Bandyopadhyay, Hyder, Mahinder, Anantha, Liu, and Gondala]{pusateri2024retrieval}
Ernest Pusateri, Anmol Walia, Anirudh Kashi, Bortik Bandyopadhyay, Nadia Hyder, Sayantan Mahinder, Raviteja Anantha, Daben Liu, and Sashank Gondala.
\newblock Retrieval augmented correction of named entity speech recognition errors.
\newblock \emph{arXiv preprint arXiv:2409.06062}, 2024.

\bibitem[Robatian et~al.(2025)Robatian, Hajipour, Peyghan, Rajabi, Amini, Ghaemmaghami, and Gholampour]{robatian2025gec}
Amin Robatian, Mohammad Hajipour, Mohammad~Reza Peyghan, Fatemeh Rajabi, Sajjad Amini, Shahrokh Ghaemmaghami, and Iman Gholampour.
\newblock Gec-rag: Improving generative error correction via retrieval-augmented generation for automatic speech recognition systems.
\newblock \emph{arXiv preprint arXiv:2501.10734}, 2025.

\end{thebibliography}

\end{document}